# An Internal Clock Based Space-time Neural Network for Motion Speed Recognition

Junwen Luo†
Computing Technology Lab
Alibaba Group
Shanghai, China
junwen.luo@alibaba-inc.com

Jiaoyan Chen
Computing Technology Lab
Alibaba Group
Shanghai, China
yanqie.cjy@alibaba-inc.com

## ABSTRACT

In this work we present a novel internal clock based space-time neural network for motion speed recognition. The developed system has a spike train encoder, a Spiking Neural Network (SNN) with internal clocking behaviours, a pattern transformation block and a Network Dynamic Dependent Plasticity (NDDP) learning block. The core principle is that the developed SNN will automatically tune its network pattern frequency (internal clock frequency) to recognize human motions in a speed domain. We employed both cartoons and real-world videos as training benchmarks, results demonstrate that our system can not only recognize motions with considerable speed differences (e.g. run, walk, jump, wonder(think) and standstill), but also motions with subtle speed gaps such as run and fast walk. The inference accuracy can be up to 83.3% (cartoon videos) and 75% (real-world videos). Meanwhile, the system only requires six video datasets in the learning stage and with up to 42 training trials. Hardware performance estimation indicates that the training time is 0.84-4.35s and power consumption is 33.26-201mW (based on an ARM Cortex M4 processor). Therefore, our system takes unique learning advantages of the requirement of the small dataset, quick learning and low power performance, which shows great potentials for edge or scalable AI based applications.

## 1 Introduction

Nowadays Artificial Neural Networks (ANNs)[1] achieve huge successes and become one of the key factors leading to the next generation industrial revolution. And it is a game-changing player in some industrial fields such as face recognition[2], auto-driving and natural language processing[3]. It progresses rapidly and meanwhile, it suffers several main constraints such as requirements of a large amount of training data, low fault tolerances and without cognitive computing functions[4]. This is fundamentally different from how our brains process information[5], and these issues are not solved yet. Therefore, there is a small portion of researchers follow the other path and try to overcome this dilemma: Spiking Neural Networks (SNNs) come of the age[6] and use temporal-spatial based processing and event-driven mechanisms[7][8]. And the core principles of SNNs are to replicate fascinate brain computing behaviours[9][10]: ultra-low power consumption, self-learning and strong fault tolerances. Unfortunately, up to now there is still a considerable gap between ANNs and SNNs regarding the application levels. Based on our limited knowledge, we conclude several issues as below:

- ***Lack efficient SNN training algorithms***

The mainstream SNN training algorithms such as Spiking-timing dependent plasticity (STDP) are widely used in the neuromorphic computing field. For example, ODIN [11] develops a 10-neuron SNN and employed SDSP learning algorithm for MINST dataset testing, the system demonstrates its capability with 84.5% classification accuracy. Meanwhile, [12][13][14] shows similar results by using SNNs based STDP learning algorithms. However, STDP is a local training algorithm which strongly limits its application. Also, there is a large number of groups investigate SNNs based backpropagation or gradient descent algorithms which similar to ANNs training framework[15][16]. However, these kinds of algorithms seem feeble and don't fit SNNs nature computing features.

- ***Mimicking a brain from an obscure level***

Simulation of a brain computing can be either from a high bio-plausible level Hodgkin-Huxley neuron model[17] or a high mathematical level leakage-and-integration neuron model[18]. Similarly at a network level, modelling of a small neural network can perform plasticity, adaption and compensation [19][20][21], while formulating a large scale network (100,000) takes advantage of cognitive computing features[22][23]. We are confused about at which level the neuromorphic system should learn from a brain. The obvious reason is the brain is not fully understood yet[24], and more importantly, neuromorphic engineers are not well recognized this point when they develop systems. As a result of this, the developed system doesn't reflect SNN computing features properly.

- ***Bottom-up approach is not enough for SNNs applications***

Currently neuromorphic computing fields are largely focused on hardware architecture design such as Neurogrid[25], TrueNorth[26] and neural processors[27]. They all made a significant contribution on this field and demonstrate the capabilities to simulate either a million neurons or complicated ion channel mechanisms in real-time. One potential risk of this bottom-up approach is that the emerging algorithms may not well fit into developed hardware, and results of no killer applications. The algorithm, software, hardware, and application should be fully taken into accounts when we design a neuromorphic computing system.

Therefore, by considering these factors above and inspired by the biological cerebellum Passenger-of-Timing (POT) mechanism[28][29], we propose a novel SNN based learning system for speed recognitions. As it is shown in Figure. 1, the system consists of a spike train encoder, an internal clock based SNN, a pattern transformation block and a Network-Dynamic Dependent Plasticity (NDDP) learning block. The main principle is that motion speed can be differentiated via a trained SNN internal clock timing information. By applying both cartoon and real-world videos, results demonstrate that under a constrained hardware resources environment, the proposed system can not only recognize motions with considerable speed differences (*e.g.* run, walk, jump, wonder and standstill) but also motions with subtle speed gaps such as slow run and fast walk. Therefore, the key contributions are as followed:

- *Algorithms level*: developed a novel SNN training algorithm from a global network dynamic perspective which can reflect SNN key computing advantages: requirement small datasets (6 videos in our work); quick learning (6-40 training trails) and has certain cognitive computing behaviours (can differentiate real-world videos based on trained cartoon videos).

- *Applications level:* the proposed system can be applied on IoT fields for speed recognition due to its ultra-low power consumption(33.26mW), short-latency (0.84s) and usage of limited hardware resources (can be implemented on a typical ARM Cortex M4 controller). And this will enable system learning capabilities on edges or end devices.

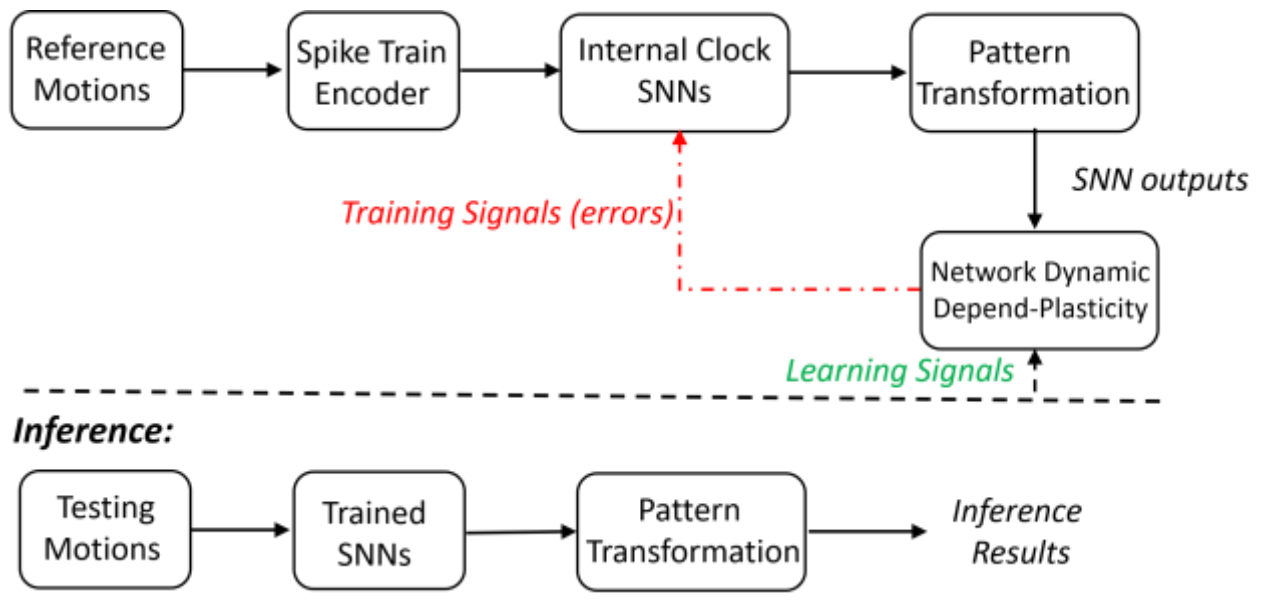

**Figure 1: the internal clock based SNNs learning system.**

## 2 The learning system

An internal clock based SNN learning system has three stages for training and learning: 1)*information translation*: the input motion videos are transformed into spike trains via a spike train encoder; 2)*training*: by given pre-defined learning signals, the SNN modify its global dynamic pattern frequency (internal clock frequency) via NDDP learning rules to minimize errors (cost function); 3)*inference*: the trained SNN differentiates input motions based on mean firing rates. The detailed individual blocks are described in Figure 2.

### 2.1 The spike train encoder

A temporal-spatial spike train encoder aims to reduce redundant information both in time and space domain, and only events related information is given into SNNs. The equation is as below:

$$S = \sum_{i=1}^{f} \sum_{j=1}^{n/\Delta s} [A_j^i - A_j^{i-\Delta t}]^+ \tag{1}$$

Where $S$ is the total information (bits) given to the neural network, $f$ is an input video frame number, $n$ is a network neuron number (pixel number), $\Delta s$ is a spatial resolution that converts several pixel values into a single one, $\Delta t$ is a differential timing between a current frame and reference frame. $A_j^i$ is pixel $j$ at frame $i$ activities: A = 1 indicates spiking, otherwise A = 0 (a function $[u]^+$ equals 1 when $u \cong 0$, otherwise equals 0). As in Figure 2 (a-b) displays, the reference video motion is converted into spike trains, video frames are encoded into corresponding neuron spike trains as inputs. Figure 2(b) displays a detailed example of converting *run* motion video into spike trains in a contour plot format.

### 2.2 The internal clock based SNN

Based on the previous work[29], we develop a new spiking neural network and with two types of inputs: one is synaptic inputs of excitatory and recurrent inhibitory inputs from the other neurons, and the other one is from motions spike trains. The model is tailor modified leaky integrate-and-fire model as equation shown below:

$$u(t) = [I - \sum_{j=1}^{N} w_{ij} \sum_{s=1}^{t} \exp\left(-\frac{t-s}{\tau}\right) A_j(s-1)]^+ \tag{2}$$

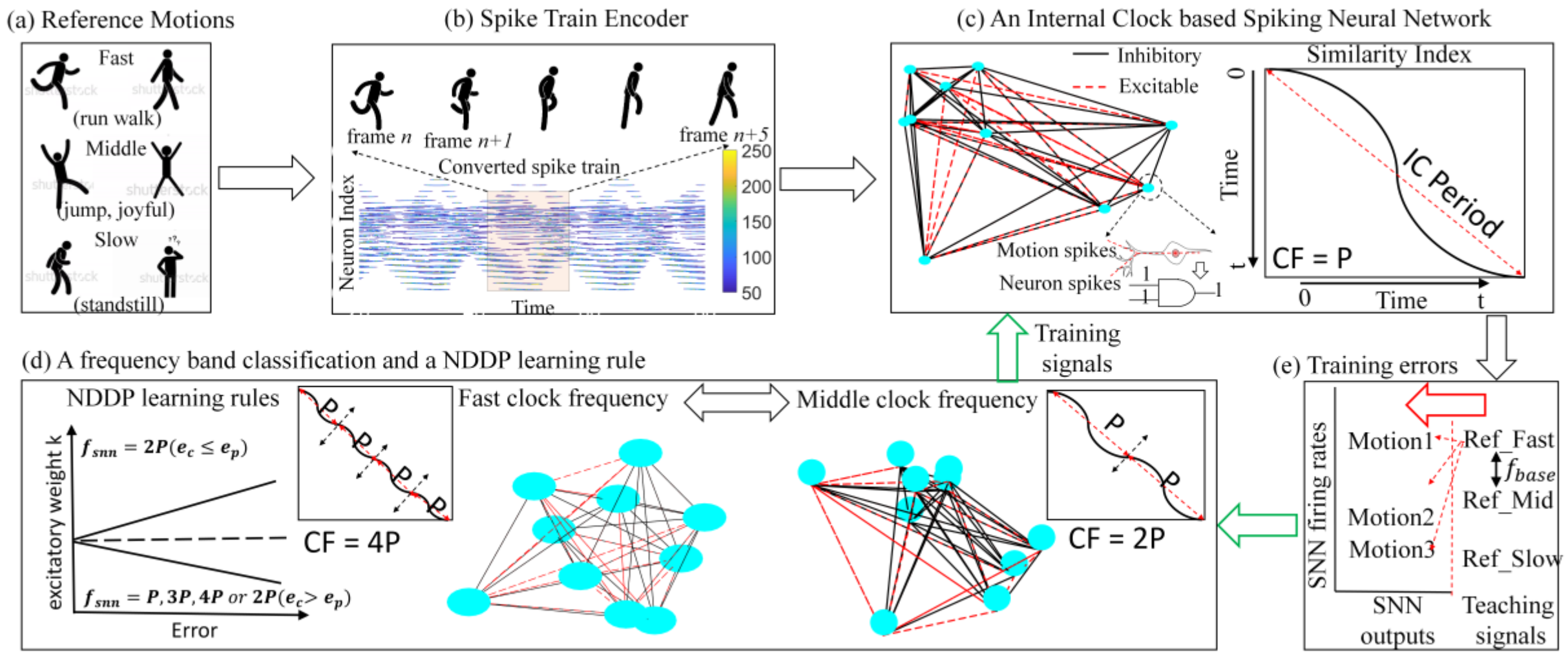

**Figure 2: the internal clock based SNNs learning system. (a) is a library of reference motion video: *run*, *fast walk*, *slow walk*, *jump*, *joyful* and *standstill*; (b) is a spike train encoder example of converting a *run* video into spike trains (contour plot); (c) an internal clock based on SNN: the cycle indicates neuron module, black lines are inhibitory synapses and red lines are excitable synapses. The neuron computing mechanisms are also shown at the bottom right. The internal clock is calculated by a similarity index; (d) is a frequency band classification and an NDDP learning rule and (e) is a training error description.**

Where $u(t)$ and $A_j$ is neuron membrane potentials and activity states; $I$ is an external afferent input signal and $w_{ij}$ represents neuron j to neuron i. A function $[u]^+$ equals 1 when $u \cong 0$, otherwise equals 0. The final SNN outputs are the results of Boolean AND logic operation between neuron spikes $u(t)$ and motion spikes $m_t^i$, where $m_t^i$ is a motion spike index $i$ at timing $t$ activities (1 or 0). This is to build a correlation between internal SNNs and external world dynamics (Figure 3a). Neuron model also has long temporal integration of activities of neurons. This is described by the summation with respect to $s$, $\tau$ is the decay time constant. The neural network global dynamic pattern frequency (internal clock frequency) is described by using the similarity index, which is shown at equation (3):

$$C(t_1, t_2) = \frac{\sum_{i=1}^{N} z_i(t_1) z_i(t_2)}{\sqrt{\sum_{i=1}^{N} z_i^2(t_1)}\sqrt{\sum_{i=1}^{N} z_i^2(t_2)}} \quad (3)$$

Where $C(t_1, t_2)$ equals 1 if the activity pattern $z_i(t_1)$ and $z_i(t_2)$ are identical, and it equals 0 if they are orthogonal, which illustrates that there is no overlap. $t_1$ and $t_1$ are simulation time index from 0 to the last simulation step. As Figure 2(c) shows, the internal clock frequency can be calculated by evaluating repetitive pattern frequencies (internal clock frequency). Here we employ above similarity index to measure repetitive pattern frequencies.

## 2.3 The network dynamic dependent plasticity learning rule

The system learning process is divided into two stages: *1) STAGE1*: frequency band classifications; *2) STAGE2*: an NDDP training. Before the training process, a teaching signal is given to describe dataset video motion frequency (*e.g.* from fast to slow):

$$\varepsilon = \{(m_0; f_0; w_0\}, (m_1; f_1; w_1\}, \dots, (m_i; f_i; w_i\}, \dots\}; \quad (4)$$

$$f_0 > f_1 > \cdots f_i > \cdots$$

$$f_i - f_{i+1} > f_{base}$$

Where $m$ is video motion index and $f, w$ is its motion frequency (*e.g.* walking frequency) and rank weights. Motion videos will be ranked from high to low based on frequencies (these are calculated based on video information). A variable $f_{base}$ is given to distinguish different motion types.

At frequency band classification stage (this stage purpose is to significantly reduce training time), an SNN will be configured into the four different internal clock frequencies sequentially: $f_{snn} = P$ (slow, patterns are no overlap at Figure2 (b-c)); $f_{snn} = 2P$ (middle, network patterns overlap twice at Figure 2(b-c) ); $f_{snn} = 3P$ (fast, network patterns overlap more than two times at Figure 2(b-c) left) and $f_{snn} = 4P$ (ultra-fast, network patterns overlap in most of the time at Figure 2(b-c)). The internal clock frequency modification is achieved in equation (5):

$$f_{snn} = F(b, \tau, k) \quad (5)$$

Where $b$ is a neuron modular size, which means how many neurons share the same synaptic connections; $\tau$ is the neuron model decay time constant and $k$ is a network excitatory synapse weight. Then each motion spike train will be sent into SNNs for calculating the mean firing rates. The outputs will be ranked in the order identical to the teaching signal sequence (from high to low). The training errors are calculated as below (Figure 2(e)):

$$e = \sum_{i=0}^{n}(w_i - w_i^s) + \sum_{j=0}^{m}\left[f_j^s - f_{j+1}^s\right]^+ \quad (6)$$

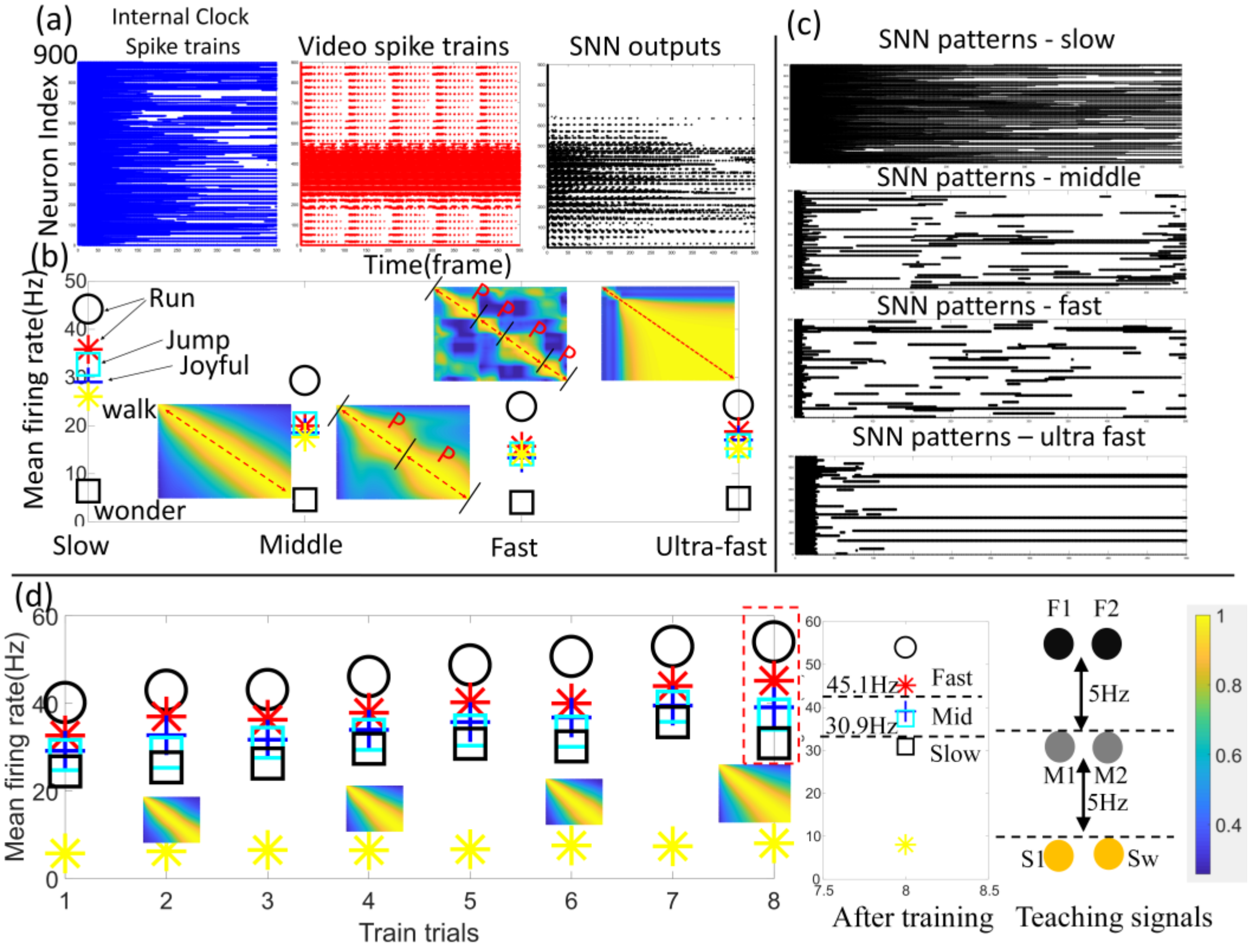

**Figure 3: The system training process of motion recognition: (a) is an SNN computing example. The left one is an internal clock based SNN spiking patterns. The middle one is a motion video based spiking patterns. The right one is a final SNN spike pattern outputs.; (b) is a result of frequency band classification; (c) is a result of SNNs spiking burst pattern under different internal clock frequency; (d) is an NDDP training results. The training trial number is 8 and $f_{base}$ is 5Hz.**

Where $w_i^s$ and $f_i^s$ is the rank weight and mean firing rates of the $i_{th}$ motion videos. Function $[u]^+$ equals 1 if $u < f_{base}$, otherwise equals 0. The SNN frequency band with minimal errors $e$ will be selected for the next stage training.

At an NDDP training stage, as it is shown in Figure 2(d), the selected SNN global excitatory synaptic weight $k$ will be tuned based on training errors $e$. This is achieved by equation (7):

$$\begin{cases} k = k(1-\delta), & (e_c > e_p) \\ k = k(1+\delta), & (f_{snn} = 2P; e_c \le e_p) \end{cases} \quad (7)$$

Where $\delta$ is a training rate, $\boldsymbol{e_c}$ and $\boldsymbol{e_p}$ is the current trial training errors and previous trial training errors. At each training trial, the SNN synaptic weight $\boldsymbol{k}$ will be fine-tuned until the training error equals 0. The global synaptic weight $k$ upper limit is 2.5.

## 3 Results

Three different benchmarks are tested to prove system functionalities: 1) recognition of motions with considerable speed differences: *run*, *walk*, *joyful*, *jump*, *slow walk* and *wonder*; 2) recognition of motions with subtle speed gap such as *slow run* and *fast walk*; 3) recognition of real-world motion videos based on knowledge learned from cartoon videos.

Regarding experimental setup, the neuron number N = 900 and with stimulation time T = 500 ms. Excitatory and inhibitory synapses (weight is 0) number ratio is followed binomial distribution P = 0.5. Cartoon motion videos format is 'RGB24', the resolutions are 596 by 336 pixels, the frame rate is 30 and bits per pixel is 8. Based on the developed system, spike train encoder parameters are setup as $\Delta s = 4$ and $\Delta t = 1$. For each internal clock frequency band, the parameters are configured as below: slow frequency $\{f_{snn} = P;\ b = 1, \tau = 100,\ k = 1)$; middle frequency $\{f_{snn} = 2P;\ b = 5, \tau = 100,\ k = 1)$; fast frequency $\{f_{snn} = 3P;\ b = 10, \tau = 50,\ k = 2.5)$ and ultra-fast frequency $\{f_{snn} = 4P;\ b = 15, \tau = 50,\ k = 2.5)$; the training parameter $\delta = 0.001$.

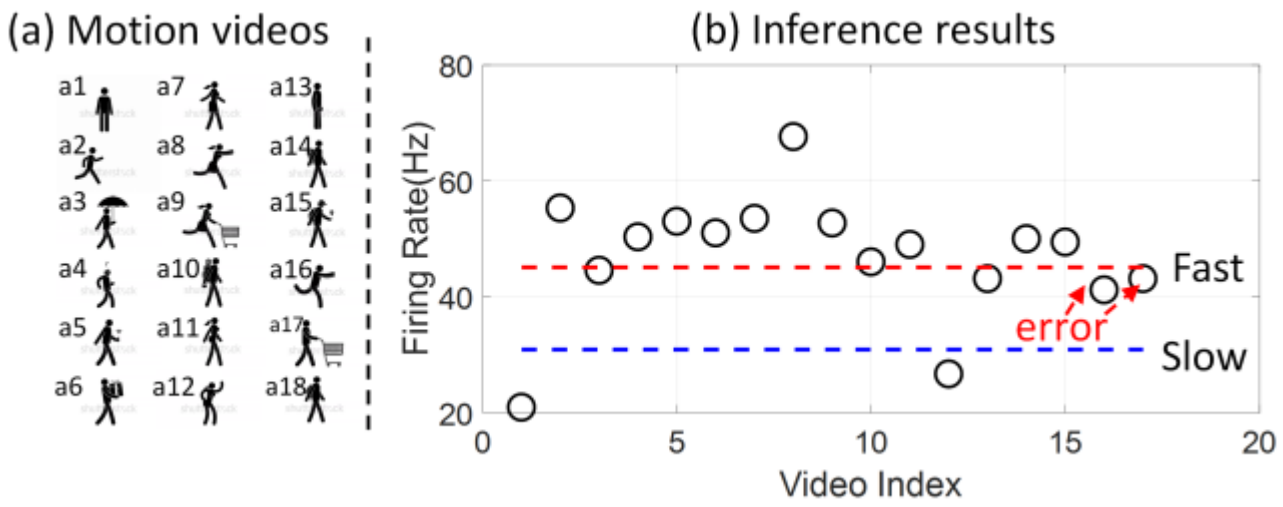

**Figure 4: Inference results of 18 motion videos. The motion video is a1) standstill; a2) people is running on the left; a3)walk with umbrella; a4) walk and listen music, a5)walk with cellphone, a6)walk with gift box; a7)walk and jump(girl); a8) run(girl); a9) run with trolley (girl); a10) walk with oxygen hose; a11)walk(girl); a12) dance(girl); a13) standstill; a14) walk with bag; a15) walk with bag and cellphone; a16) run; a17) walk with trolley and a18)walk with bag. And on the right figure are inference results.**

## 3.1 The motion recognition of *run, walk, joyful, jump, slow walk* and *wonder*

Six motion videos labelled as *run*, *walk*, *joyful*, *jump*, *slow walk* and *wonder* are served as a training dataset. The teaching signals are defined in the box below.

$$\varepsilon = \{(m_r; f_0 = 15Hz; w_r = 6)\}, \{(m_w; f_w = 15Hz; w_w = 6)\}, \{(m_{ju}; f_{ju} = 10Hz; w_{ju} = 3)\}, \{(m_{joy}; f_{joy} = 10Hz; w_{joy} = 3)\}, \{(m_s; f_s = 5Hz; w_s = 1)\}, \{(m_{wo}; f_{wo} = 5Hz; w_{wo} = 1)\}\}; \quad f_{base} = 5Hz ;$$

There are three different types of motions: fast motions with 15Hz, middle motions with 10Hz and slow motions with 5Hz. The training results are shown in Figure 3: (a) is an example of illustrating SNN computing mechanisms. The left figure is an SNN network pattern outputs (neuron spikes), the middle one is a spike train (motion spikes) from a running motion video. The final system output will be the AND logic operation of these two inputs in the time domain, which is shown at the right figure. (b) shows frequency band classification results, firing rates of each motion video is labelled in Figure 3(b) as well. The result demonstrates that an SNN with slow clock frequency has minimal training error equals 1. This indicates the motion sequences are identical with teaching signals, but there are some motion firing rates differences are less than $f_{base}$. Each SNNs spiking burst patterns and similarity index and are both displayed in Figure 3(b) and Figure 3(c). Figure 3(d) depicts NDDP training results, at 8th training trail, the SNN successfully differentiate 6 motion videos based on teaching signals.

At the inference stage as Figure 4 depicts, 18 videos are randomly selected from three types of motions. Based on the trained results, motions with mean firing rates above 45.1Hz (Figure 4 red dash line) are identical with fast movements, motions with firing rates between 45.1Hz and 30.9Hz are identical with medium movements and motions with firing rate below 30.9Hz (Figure 4 blue dash line) are identical with slow movements. Only two motions a17(fast) and a18(fast) are differentiated into medium motions, the overall accuracy is 88.9%.

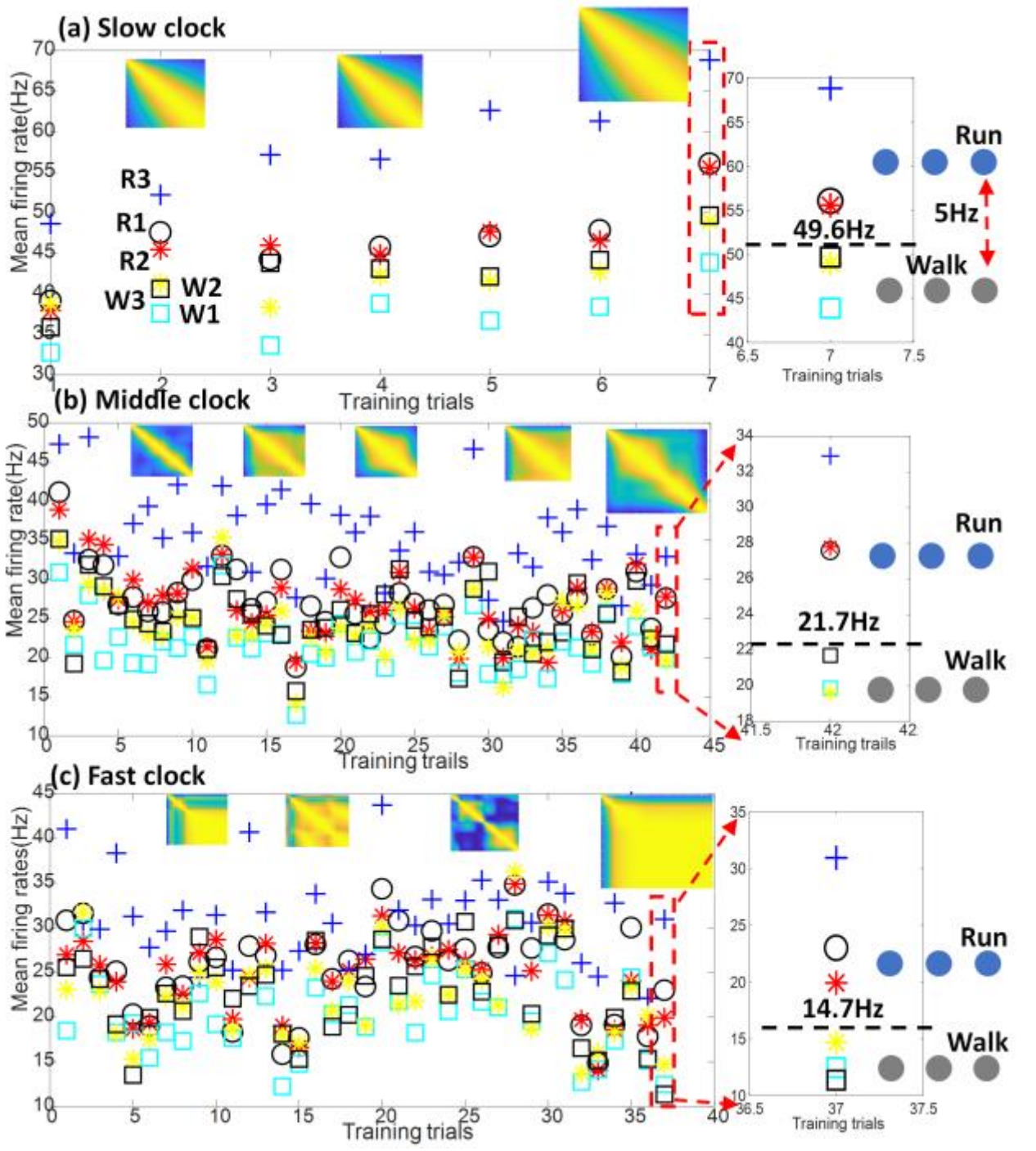

**Figure 5: Training results of motion *slow run* and *fast walk*. At a slow clock condition, the neural network requires 7 training trials; at a middle clock condition, the neural network requires 45 training trails; and at a fast clock condition, the neural network requires 35 training trails.**

## 3.2 The motion recognition of *slow run* and *fast walk*

In order to further prove developed system capabilities, six videos {slow run and fast walk} with tiny frequency gaps are chosen in this experiment. In this case teaching signals are defined in the box as below:

$$\varepsilon = \{(m_{r1}; f_{r1} = 15Hz; w_{r1} = 6)\}, \{(m_{r2}; f_{r2} = 15Hz; w_{r2} = 6)\}, \{(m_{r3}; f_{r3} = 15Hz; w_{r3} = 6)\}, \{(m_{w1}; f_{w1} = 10Hz; w_{w1} = 3)\}, \{(m_{w2}; f_{w2} = 10Hz; w_{w2} = 3)\}, , \{(m_{w3}; f_{w3} = 5Hz; w_{w3} = 3\}\}; \quad f_{base} = 5Hz ;$$

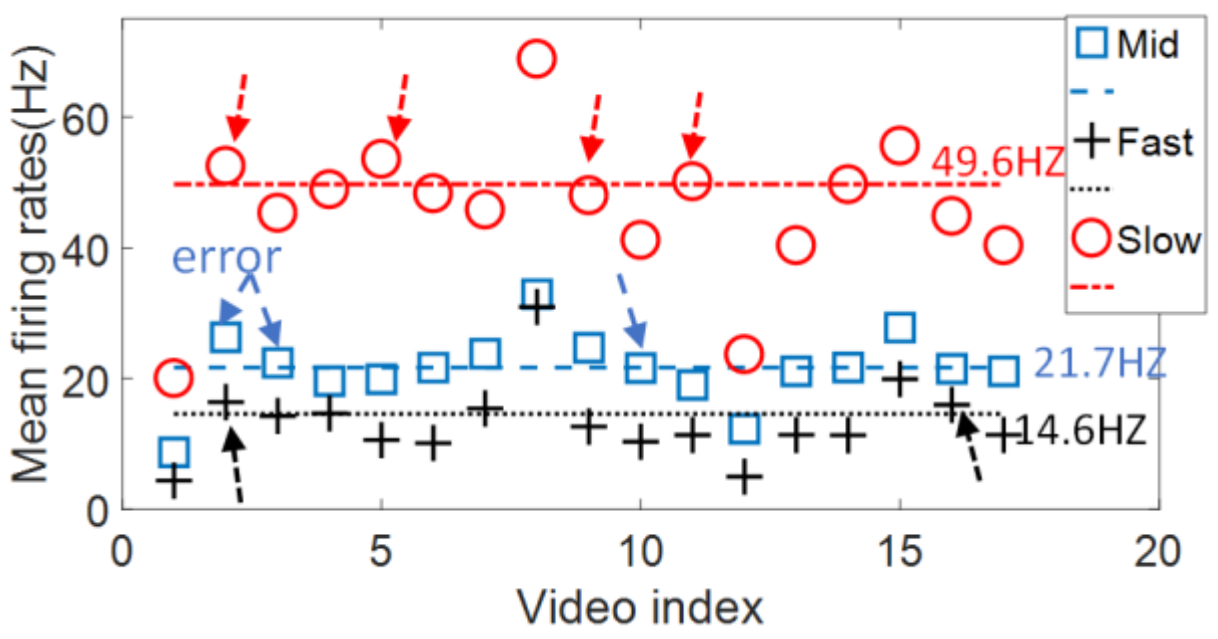

**Figure 6: Inference results of cartoon motion *slow run* and *fast walk*. The results of a slow clock condition is labeled in red circle; the results of a fast clock condition is labeled in black cross, and the results of a middle clock condition is labeled in blue square.**

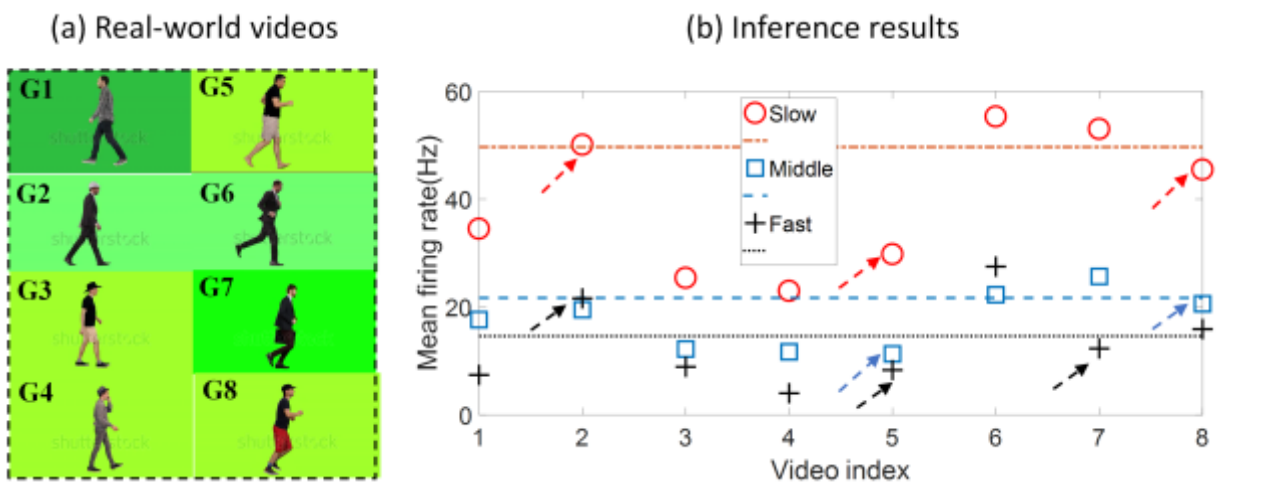

**Figure 7: Inference results of real-world motion *slow run* and *fast walk*. The results of a slow clock condition is labeled in red circle; the results of a fast clock condition is labeled in black cross, and the results of a middle clock condition is labeled in blue square.**

At first frequency band classification stage, SNNs with three internal clock frequencies {slow, middle, fast} have the same training errors. Therefore, NDDP training is all applied in each frequency domain at the second stage. As Figure 5 depicts, an SNN with slow clock frequency finished training at 7th trail with k = 0.64, the classification firing rate is 49.6Hz. An SNN with middle clock frequency finished training at 43rd trail with k = 1.2685, the classification firing rate is 21.7Hz. and an SNN with fast clock frequency finished training at 36th trail with k = 1.732, the classification firing rate is 14.7Hz.

Inference results are displayed at Figure 6, an SNN with slow clock frequency has 4 errors (red arrows at video index 3,6,8,12) with accuracy 77.8%; an SNN with middle clock frequency has 3 errors (blue arrows at video index 3,4,10) with accuracy 83.3%; and an SNN with fast clock frequency has 2 errors (black arrows at video index 3,17) with accuracy 88.9%. The results are summarized in Table1.

## 3.3 The motion recognition of real-world videos

**Table 1: System performance and hardware implementation estimation**

| SNN Clock frequency | Training trails | Accuracy Cartoon | Accuracy Real-world | Latency (T/I) | Power consumption |
|---|---|---|---|---|---|
| Slow | 7 | 77.8% | 62.5% | 0.84s/0.08s | 33.26mW |
| Middle | 42 | 83.3% | 75% | 5.08s/0.08s | 201mW |
| Fast | 36 | 88.9% | 62.5% | 4.35s/0.08s | 172.2mW |

Since the real-world motion such as walk and run share the identical repetitive spike train patterns with cartoon videos, we did an inference for real-world motion videos (slow run and fast walk) based on cartoon videos trained SNNs. The results are displayed at Figure. 7, 4 videos with a fast walk (G1-G4) and 4 videos with a slow walk (G5-G8) are employed for at this experiment. The system with slow clock frequency has 3 errors (red arrows at video index 2,5,8) with accuracy 62.5%; the system with middle clock frequency has 2 errors (red arrows at video index 5,8) with accuracy 75%, and the system with fast clock frequency has 3 errors (red arrows at video index 2,5,7) with accuracy 62.5%. The results are summarized in Table 1.

## 3.4 The estimation of hardware implementation performance

We estimated algorithms hardware implementation results on our previously designed embedded-ASIC hardware[30][31]. For a single training trail the latency is 0.08s, hence the total training time for SNNs with slow, middle fast clock frequency is 0.84s, 5.08s and 4.35s. And power consumptions of each case are 33.26mW, 201mW and 172.2mW. Here an event-driven implementation technique is not applied here so the total power can be further optimized in the near future.

# 4 Discussions

## 4.1 The model advances

In this work we develop a novel internal clock based SNN learning system for speed recognition. The system key advances are as below:

- ***Requirement of a small training dataset***

The developed system employs 6 motion videos for training purposes, and inferences 18 motion videos. The ratio of training and inference dataset is 1:3. The key reason is that designed SNN captures the common speed properties of motion videos and transforms them into a spiking-burst pattern domain for processing.

- ***Quick learning performances***

For SNNs with slow clock frequencies, only less than 10 training trails are required, while for SNNs with middle and fast clock frequencies, the training trail number is up to 50 times. This is due to we modify neural network global spiking-patterns rather than individual neurons. Based on our previous hardware implementation work[32][31], we estimated the latency on a typical ARM Cortex M4 processor is less than 6 seconds for 50 times training. The details are summarized in Table 1.

- ***Has certain cognitive behaviors***

By using cartoon videos trained SNN, the system can also differentiate real-world run and walk videos with certain accuracies. This proves that the system has basic cognitive learning behaviors in a spiking-pattern domain.

- ***The SNN with specific behaviors***

Inspired by the work[33], the developed SNN has tailor-designed internal clock timing behaviors[34] at initial stages. This will strongly beneficial to one-shot /few learning performances.

## 4.2 Model Applications

One of the most promising applications for the developed algorithm is the edge/IoT fields since developed system hardware implementation only has less than seconds latency and 33-201mW power consumption, the typical IoT based embedded processors can easily implement developed algorithms and enable learning behaviors at the end device level.

## 4.3 Model limitations and future work

Currently developed SNNs fully focus on timing representation via internal clocking behaviors. However, in some special cases, large dynamic events on the spatial domain can also exert vital effects on inference results such as video index 3 (walk with a big umbrella). Compare to the other work[35], the developed NDDP results of variable training time and uncertain results. Also, the maximum movement speed that can be recognized by the developed SNN is still required further explorations. This is closely related to the large-scale datasets and the developed NDDP rule. In the next stage, we will investigate introducing spatial domain information representation mechanisms[36] and introduce standard dataset videos[37] for the training as well as algorithm optimizations.

# ACKNOWLEDGMENTS

We would thank for the great supports from computing technology lab team members.